# Speech and Text-Based Emotion Recognizer

Varun Sharma

National University of Singapore

sharmavarun.s@u.nus.edu

*Abstract*— Affective computing is a field of study that focuses on developing systems and technologies that can understand, interpret, and respond to human emotions. Speech Emotion Recognition (SER), in particular, has got a lot of attention from researchers in the recent past. However, in many cases, the publicly available datasets, used for training and evaluation, are scarce and imbalanced across the emotion labels. In this work, we focused on building a balanced corpus from these publicly available datasets by combining these datasets as well as employing various speech data augmentation techniques. Furthermore, we experimented with different architectures for speech emotion recognition. Our best system, a multi-modal speech, and text-based model, provides a performance of UA(Unweighed Accuracy) + WA (Weighed Accuracy) of 157.57 compared to the baseline algorithm performance of 119.66

## I. INTRODUCTION

Although there have been significant advancements in the field of artificial intelligence, natural interaction between humans and machines remains a challenge, partly due to the machine's inability to comprehend our emotional states. When people speak, the sounds they produce convey not only the intended linguistic message but also a wealth of additional information about themselves and their emotional states. Affective computing, a sub-field of Human-Computer Interaction (HCI) systems, aims to facilitate natural interaction with machines by direct voice interaction instead of using traditional devices as input.

Speech-based Emotion Recognition (SER) is a relatively new field that has emerged in the past few decades. The earliest research in this area can be traced back to the 1990s when scientists began to explore the possibility of using computers to recognize emotions in speech. Early efforts focused on identifying basic emotions such as happiness, sadness, anger, and fear, using simple acoustic features such as pitch and intensity. Over time, researchers developed more sophisticated techniques for extracting emotional information from speech signals, incorporating advanced machine learning algorithms and drawing on knowledge from fields such as psychology and linguistics. There have been significant advances in SER technology in recent years, thanks to the development of deep learning algorithms. Deep learning algorithms can learn complex patterns in data, and they are very effective for SER.

Prominent among these have been wav2vec 2.0 [3] and Hu-BERT [13] models developed by Facebook. Both of these models have been pre-trained on large amounts of audio data which enables them to achieve impressive performance on speech processing tasks.

In the past, the SER domain has been plagued by 2 significant challenges. First has been *"how to model emotional representations from speech signals"*. Traditional methods focus on the efficient extraction of hand-crafted features such as intensity, pitch, formants, zero crossing rate, and Mel Frequency Cepstrum Coefficients (MFCCs) to name a few. Although researchers have done a tremendous amount of work in this field, there are still issues of speech feature choice and the correct application of feature engineering that remain to be solved in the domain of SER [16].

Another obstacle has been the *"availability of large, high-quality datasets"* for speech emotion recognition (SER). While there are several publicly available SER datasets, they tend to be smaller in size and less diverse than image datasets. This can make it more challenging to train and evaluate SER models, as there may be less data available to learn from.

This paper tries to address both of these challenges. The contributions of this paper are chiefly (1) the analysis of various machine learning and deep learning architectures for emotion classification. (2) combining different publicly available datasets on SER to form a larger corpus. (3) applying speech augmentation techniques to balance and expand the corpus. (4) using transfer learning to finetune the Hubert model on this corpus. (5) evaluating the impact of jointly training audio embeddings from Hubert with the text embeddings obtained from Whisper[1] and BERT models and finally (6) using UA+WA as a metric for optimizing and evaluating the system.

The rest of the paper is structured as follows. In Section 2, we review the related work on SER. In Section 3, we describe in detail, the datasets, methods, and evaluation methods that we used. In Section 4, we present the results of our experiments and compare them against each other. In this section, we also present benchmark test results on **RAVDESS** dataset. In Section 5, we discuss the deployment and web application-based inference and in Section 6, we conclude the paper and suggest future directions.

## II. RELATED WORK

The ability to understand and generate emotions in speech has been a key area of research for many decades. Their importance has increased even more in recent times, due to the widespread use of virtual assistants (such as Siri, Alexa, or Google Home)

---

[1]https://openai.com/blog/whisper/

and their applications in the healthcare industry.

*A. Conventional SER*

Early work on SER included using manually-designed features as input, such as audio energy, zero-crossing rate, and Mel-frequency cepstrum coefficients (MFCCs) [21]. Different types of generative models were used as classifiers, such as Gaussian mixture models (GMMs) [38], hidden Markov models (HMMs) [26], and support vector machines (SVMs) that use a discriminative approach [35].

Many basic acoustic parameters related to prosody and spectrum of speech, such as pitch, formant frequencies, voice quality measures, speech energy, and speech rate, were shown to be associated with emotional intensity and emotional processes [31] [32], [1]. Ververidis and Kotropoulos explored the use of more advanced parameters such as the MFCCs, spectral roll-off [37]. Furthermore, TEO features [34][12], spectrograms [29] also achieved good SER performance.

But, finding the most suitable acoustic features that can distinguish different emotions has been a major but also difficult challenge of SER. The research progress was slow and showed some discrepancies among studies. Therefore, the research focus shifted to methods that do not require or reduce the need for prior knowledge of the best features and replace it with automatic feature generation methods provided by neural networks.

*B. Deep learning-based SER*

Motivated by the remarkable progress in computer vision, recent studies [30][2][42][27] have made significant improvements on SER by treating spectral features as images.

Fayek et al [10] ,investigated SER from short frames of speech spectrograms using a DNN. An average accuracy of 60.53 (six emotions eINTERFACE database) and 59.7 (seven emotions—SAVEEdatabase) was achieved. A similar but improved approach led to 64.78 of average accuracy (IEMOCAP data with five classes)[9].

The simplest DNN systems for emotion recognition are feedforward networks that are built on top of the utterance level feature representations [11]. Recurrent Neural Networks (RNN) [17] are a class of neural networks that have cyclic connections between nodes in the same layer. These net- works capture the inherent temporal context in emotions and have shown improved performance for classification task [20]. Another class of DNNs, Convolutional Neural Nets- works (CNN) [19], capture locally present context, and patterns, working on frame-level features. CNNs enable the training of end-to-end systems where the feature representations and classification are trained together using a single optimization. Few works have analyzed the performance of CNNs for speech emotion classification [23], [2]. Cummins et al. [8] further built image-based CNNs on spectrogram features.

Recently, attention-based models have made significant progress on SER[22][43] . For example, multi-head attention maps are learned by convolutional operations to select important information according to surrounding information [41]. Moreover, area attention is further introduced to compute the importance from different ranges of convolutions [40].

*C. Transfer Learning*

Transfer learning is a technique that allows a model to leverage the knowledge learned from a source domain and apply it to a target domain. For example, if you have a model that can recognize cars in images, you can use some of the features learned by that model to help you recognize trucks in images.[2]

The artificial intelligence (AI) field is undergoing a major shift in recent years, moving from specialized architectures that are designed for a single task to general-purpose foundation models that can be easily transferred to different use-cases [4]. These foundation models are often trained on large datasets, using proxy tasks to avoid the need for hard-to-acquire labels, and then fine-tuned on (small) sets of labeled data for their target tasks. This transfer learning technique has been very effective in computer vision [7], NLP [36], and computer audition [3] – including SER [39][28]. Among others, wav2vec 2.0 [3] and HuBERT [13] have emerged as promising foundation models for speech-related applications. They have been successfully transferred to (mostly categorical) SER by previous works.

HuBERT shares its architecture with wav2vec 2.0, but it diverges in its training approach. In place of a contrastive loss, HuBERT adopts an offline clustering technique to generate noisy labels for pre-training a Masked Language Model. It processes masked continuous speech features and forecasts designated cluster assignments. The predictive loss is enforced within the masked segments, compelling the model to develop sophisticated representations of unmasked inputs to effectively deduce the intended outcomes of the masked elements.

In this work, we evaluate several approaches to model the SER problem ranging from baseline machine learning approaches to CNNs, to finally fine-tuning Hubert. We also explore augmenting the speech embeddings with text transcription embeddings.

### III. METHODS

After a thorough literature review, an iterative modeling approach was employed for developing the solution.The following sections discuss each step in detail:

*A. Dataset curation*

**Concatenation:** The first step was to curate a robust dataset. A common problem seen in the SER domain has been the lack of quality datasets. There are several publicly available datasets for SER, however, they are still limited when compared to other domains like Image and Text. Another issue is the lack of a standard set of labels for human emotions - different datasets use slightly different sets of emotions. Hence, it was decided to combine and standardize 5 publicly available datasets to form a single large corpus.

---

[2]Wikipedia

A brief description of each of these datasets is given below:

1. CERMA-D: The CREMA-D dataset[6] is a publicly available dataset for Speech Emotion Recognition (SER). It stands for the **Crowd-sourced Emotional Multimodal Actors Dataset** and contains audio and video recordings of 91 actors, aged 20 to 74, speaking 12 sentences in six different emotions: anger, disgust, fear, happiness, sadness, and neutral. The actors were recorded in a studio setting and the recordings were annotated by multiple annotators to ensure the accuracy of the emotion labels.

2. SAVEE: The SAVEE dataset[15], which stands for **Surrey Audio-Visual Expressed Emotion** contains recordings from four native English male speakers, who were postgraduate students and researchers at the University of Surrey aged from 27 to 31 years. The emotions in the dataset are described in discrete categories: anger, disgust, fear, happiness, sadness, surprise, and neutral. The text material consisted of 15 TIMIT sentences per emotion: 3 common, 2 emotion-specific, and 10 generic sentences that were different for each emotion and phonetically balanced. This resulted in a total of 120 utterances per speaker.

3. TESS: The TESS dataset[33], which stands for **Toronto Emotional Speech Set** contains audio recordings of two actresses, aged 26 and 64 years, speaking 200 target words in the carrier phrase "Say the word .." portraying each of seven emotions: anger, disgust, fear, happiness, pleasant surprise, sadness, and neutral. There are a total of 2800 data points (audio files) in the dataset.

4. IEMOCAP: The IEMOCAP dataset[5], which stands for **Interactive Emotional Dyadic Motion Capture** contains approximately 12 hours of audiovisual data, including video, speech, motion capture of face, and text transcriptions. The dataset was collected from 10 actors (5 male and 5 female) who performed scripted and improvised scenarios designed to elicit a range of emotions. The emotions in the dataset are described in both categorical and dimensional terms, with the categorical labels including anger, happiness, excitement, sadness, frustration, fear, surprise, and neutral. The dimensional labels include valence, activation, and dominance.

5. RAVDESS: The RAVDESS dataset[24], which stands for **Ryerson Audio-Visual Database of Emotional Speech and Song**, is another dataset for Speech Emotion Recognition (SER). It contains audio and video recordings of 24 professional actors (12 male and 12 female) speaking and singing in a range of emotions. The emotions in the dataset include calm, happy, sad, angry, fearful, surprise, and disgust. Each emotion is expressed at two levels of intensity: normal and strong

For combining these datasets, the following steps were followed (Figure 1):

- From 'IEMOCAP' dataset, *'xxx','neu','oth', and 'fru'* labels were removed.

- From 'CREMA-D' dataset, *'Neutral'* label was removed.

- From 'RAVDESS' dataset, *'neutral'* label was removed.

- From 'SAVEE' dataset, *'n'* label was removed.

- From 'TESS' dataset, *'neutral'* label was removed.

- From the remaining classes in the datasets, target emotion labels were renamed to below seven classes: *calm, happy, sad, angry, fearful, disgust and surprised*

An attentive reader would have noticed that the *'neutral'* label has been removed altogether. One reason was that the *'neutral'* label is often ambiguous and can be difficult to distinguish from other emotions. For example, often, the neutral speech sample is very similar to a calm speech sample, making it difficult for a machine learning model to accurately classify the two. Another reason for removing the 'neutral' label was that it is not considered an emotion in the traditional sense. A neutral state is often considered to be the *'absence of any particular emotion'*.

**Preprocessing:** When combining multiple datasets, there may be inconsistencies and variations in the data that can affect the performance of the model. Preprocessing techniques can help address these issues and improve the quality of the combined dataset. With that rationale, following steps were carried out:

1. **Sample rate adjustment**: Since the audio files have different sample rates, we considered resampling them to a common sample rate of 16KHZ. This step ensured that the data is consistent and compatible for further processing.

2. **Noise reduction**: A noise reduction technique was applied to remove noise from the audio signals. This improved the quality of the speech signals and enhance the extraction of relevant features during spectrogram creation.

3. **Silence removal**: Removed silent or low-energy sections from the audio files. This helped reduce noise and focus on the segments that contain speech and emotional content.

4. **Short audio removal**: Removed wav files with duration less than 1 second as they may not contain enough information for accurate emotion recognition.

This resulted in a dataset with 13,236 wav files with 7 target Emotion labels(Figure 2).The details about the dataset is listed in Table 4.

The dataset consisted of various examples with different target labels. To ensure that the model could learn from a representative sample of the data, the dataset was divided into two parts using a stratified split. A stratified split is a method that preserves the proportion of each target label in both parts. For example, if the original dataset had 50% of examples with label A and 50% with label B, then the stratified split would also have the same proportions in both parts. The stratified split was done with an 80:20 ratio, meaning that 80% of the data was used for training and 20% was used for testing. The training part was used to fit the model to the data, while the testing part was used to measure

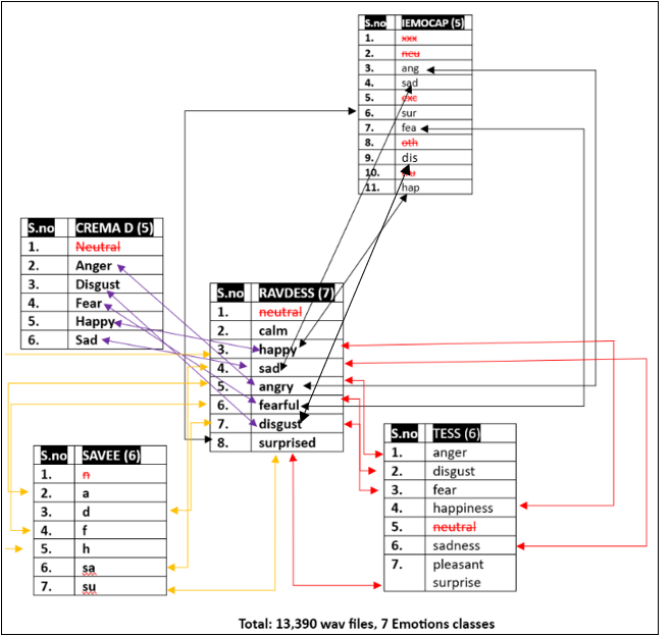

Figure 1: Combining different datasets.

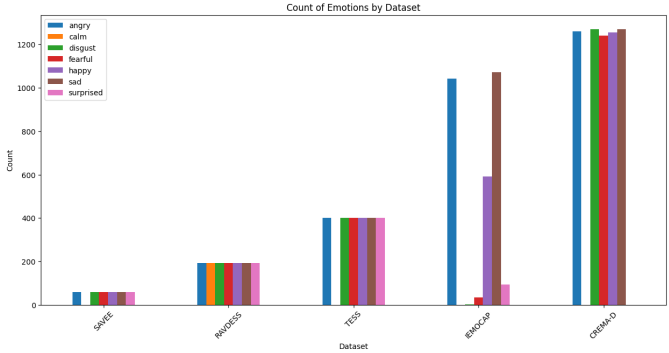

Figure 2: Combined Preprocessed Dataset Distribution

how well the model could predict new examples with unknown labels.

Furthermore, since the train dataset was imbalanced, audio augmentation techniques were applied to balance the data points across the classes:

1. **Stretching**: This technique involved changing the speed of the audio clip without affecting its pitch. Rationale was that by stretching the audio clip, the audio could be made longer or shorter, which can help the model learn to recognize emotions in speech with varying speaking rates.

2. **Pitch Shifting**: This technique involved changing the pitch of the audio clip without affecting its duration. Idea was that by shifting the pitch of the audio clip, the audio can be made higher or lower, which can help the model learn to recognize emotions in speech with varying voice pitches.

3. **Audio Gain**: This technique involved changing the volume of the audio clip. By adjusting the gain of the audio clip, one can make it louder or quieter, which can potentially help the model learn to recognize emotions in speech with varying volumes.

4. **Background Noise augmentation**: Finally, for couple of classes, background noise was added to the audio clip. Idea was to simulate real-world scenarios where speech is often accompanied by background noise.This resulted in additional samples as well as helped the model learn to recognize emotions in speech even when there is background noise present.

As described earlier , to evaluate the performance of the model, the data was split into two sets: train and test. The train set was used to train the model parameters, while the test set was used to measure how well the model generalizes to unseen data.

However, to avoid overfitting the model to the train set, a validation set was also used. The validation set was a subset of the train set that was not used for training, but for tuning the hyperparameters and selecting the best model. The validation set was obtained by splitting the train set into 80% and 20% portions. The 80% portion was used for training, while the 20% portion was used for validation. This way, the model could be evaluated on both the validation set and the test set, which were independent of each other.

Table 1,2,3 lists down the statistics of Train,validation and Test sets respectively.

*B. Baseline*

For baseline we trained a range of machine learning algorithms using a set of hand-crafted features.

**Feature Engineering**  From previous studies, emotion is found to be more correlated with energy features [14]. Hence, it was decided to use following set of features:

| Emotions | angry | calm | disgust | fearful | happy | sad | surprised | Total |
|---|---|---|---|---|---|---|---|---|
| Dataset | | | | | | | | |
| SAVEE | 38 | 0 | 52 | 54 | 38 | 37 | 108 | **327** |
| RAVDESS | 122 | 1600 | 145 | 182 | 120 | 115 | 386 | **2670** |
| TESS | 258 | 0 | 344 | 348 | 270 | 251 | 791 | **2262** |
| IEMOCAP | 691 | 0 | 0 | 20 | 379 | 707 | 201 | **1998** |
| CREMA-D | 491 | 0 | 1059 | 996 | 791 | 490 | 0 | **3827** |
| **Column Total** | **1600** | **1600** | **1600** | **1600** | **1598** | **1600** | **1486** | **11084** |

Table 1: Train dataset statistics.

| Emotions | angry | calm | disgust | fearful | happy | sad | surprised | Total |
|---|---|---|---|---|---|---|---|---|
| Dataset | | | | | | | | |
| SAVEE | 11 | 0 | 8 | 17 | 9 | 9 | 37 | **91** |
| RAVDESS | 29 | 400 | 56 | 43 | 35 | 41 | 92 | **696** |
| TESS | 67 | 0 | 70 | 77 | 55 | 69 | 202 | **540** |
| IEMOCAP | 163 | 0 | 2 | 5 | 95 | 168 | 41 | **474** |
| CREMA-D | 130 | 0 | 264 | 258 | 206 | 113 | 0 | **971** |
| **Column Total** | **400** | **400** | **400** | **400** | **400** | **400** | **372** | **2772** |

Table 2: Validation dataset statistics.

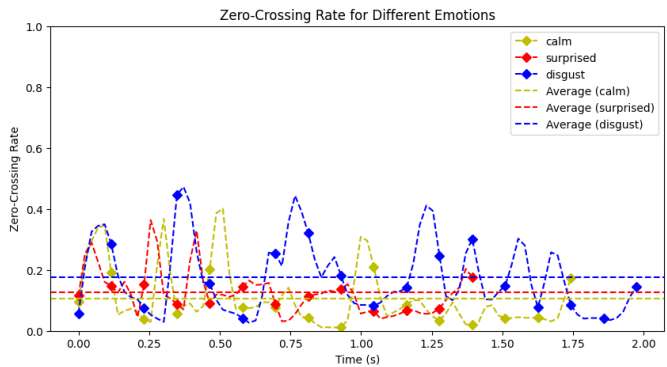

Figure 3: Zero crossing rate (number of times the signal crosses the zero line in a given timeframe) shown here is higher on average for Emotions like 'disgusting','surprised' than for 'calm' emotion.

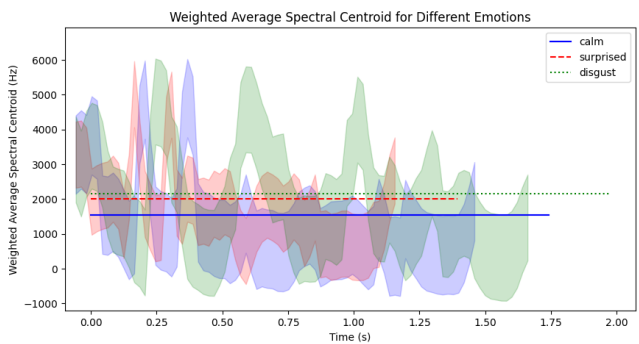

Figure 4: Spectral Centroid (measure of "centre of mass" of the spectrum), is higher for high intensity emotions like 'disgust' and 'surprised' than for emotions like 'calm'.

1. **Chromagram**: Chromagram representation maps the audio signal to the 12 pitch classes (CDEFGAB + 5 semitones). So, we got 12 features from this.

2. **MelSpectrogram**: Mel spectrogram is obtained by applying a filterbank which converts the Linear frequency scale of spectrogram to Mel scale. We use 60 mel filterbanks. So, we got 60 features

3. **MFCCs**: MFCCs are the compact representation of the spectral envelope of an audio signal derived from Mel Spectrograms. We used 20 MFCCs as another set of features.

4. **ZeroCrossingRate**: This is the rate at which a signal changes from positive to negative or vice versa. In the context of speech emotion recognition, zero cross rate can be useful for distinguishing between voiced and unvoiced sounds in an input speech signal.(Figure 3)

5. **SpectralCentroid**: The spectral centriod is a measure that indicates where the "centre of mass" of the spectrum is located. It is calculated as the weighted mean of the frequencies present in the signal, determined using a Fourier transform, with their magnitudes as the weights. In the context of speech emotion recognition, the spectral centroid can provide useful information about the distribution of energy in the frequency spectrum of a speech signal.(Figure 4)

So, a total of 94 features were extracted from the raw audio wav files. After extracting these features, standard scaling was applied to normalize the feature values.

Finally, we fitted the following machine learning models to this data to classify emotions into 7 categories:

1. KNeighbors Classifier

2. MLP Classifier

| Emotions<br>Dataset | angry | calm | disgust | fearful | happy | sad | surprised | Total |
|---|---|---|---|---|---|---|---|---|
| SAVEE | 11 | 0 | 14 | 7 | 13 | 4 | 14 | **73** |
| RAVDESS | 41 | 38 | 38 | 32 | 37 | 36 | 40 | **262** |
| TESS | 75 | 0 | 75 | 72 | 75 | 80 | 81 | **458** |
| IEMOCAP | 188 | 0 | 0 | 12 | 117 | 197 | 14 | **528** |
| CREMA-D | 276 | 0 | 258 | 262 | 259 | 272 | 0 | **1327** |
| **Column Total** | **591** | **38** | **385** | **385** | **501** | **599** | **1149** | **2648** |

Table 3: Test dataset statistics.

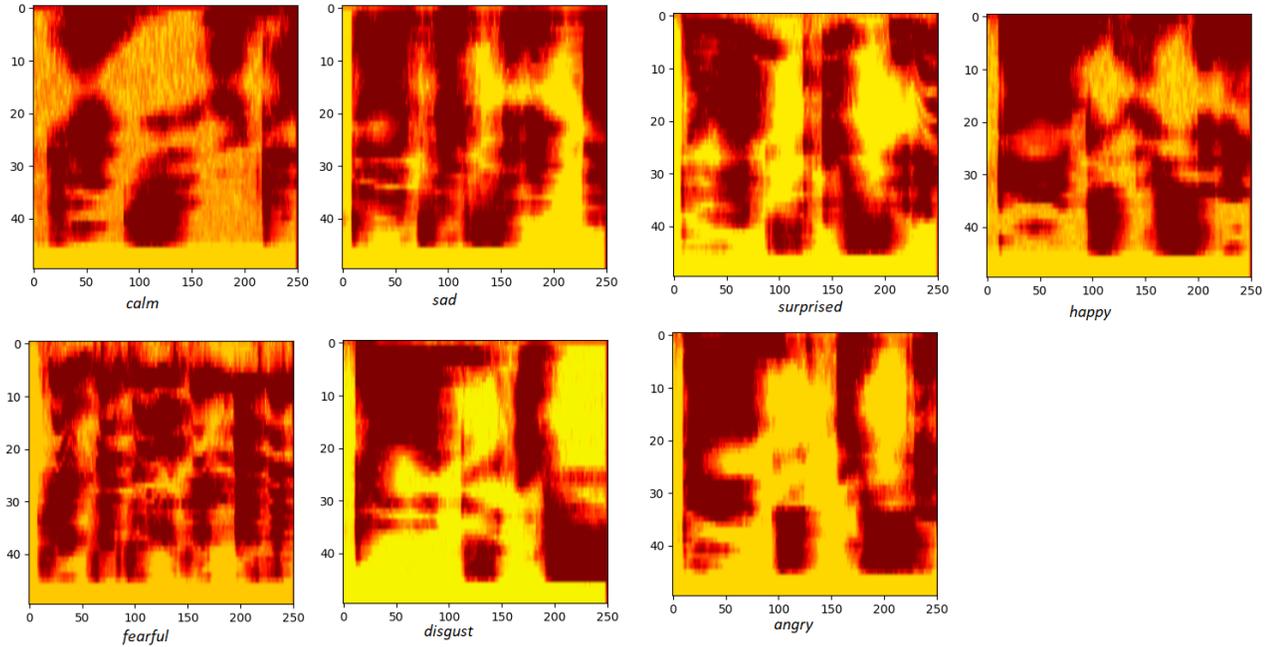

Figure 5: Spectrogram Images created out of blocks of 1-sec wav files.

| Emotions | # Utterances |
|---|---|
| angry | 2955 |
| calm | 192 |
| disgust | 1923 |
| fearful | 1927 |
| happy | 2499 |
| sad | 2995 |
| surprised | 745 |
| **Total** | **13236** |

Table 4: Combined Dataset Statistics.

3. XGB Classifier
4. LightGBM Classifier
5. Logistic Regression
6. RandomForest Classifier
7. GaussianNB
8. DecisionTree Classifier

The table 7 lists the results for the baseline test results:

With the best performing model (KNN), Grid search CV was fitted to get the optimum set of parameters. With those parameters, the model was again fitted on the same train dataset. However, accuracy did not improve further.

### C. CNN Models

Next, we experimented with different CNN architectures for solving the emotion classification problem:

1. **MobileNetV2 Model with 2D Convolution:** Similar to [18], another approach that was tried was to redefine the Speech Emotion Recognition problem as an **'image classification task'**.

To achieve this, labeled speech samples were buffered into

| Layer Type | Kernels | Kernel Size | Activation | Padding | Batch Norm | Dropout |
|---|---|---|---|---|---|---|
| Input | - | - | - | - | - | - |
| Conv1D | 256 | 8 | ReLU | Same | No | No |
| Conv1D | 256 | 8 | ReLU | Same | No | No |
| Batch Norm | - | - | - | - | Yes (After Conv1D) | No |
| Dropout | - | - | - | - | No (After Batch Norm) | 0.6 |
| Conv1D | 128 | 8 | ReLU | Same | No | No |
| Conv1D | 128 | 8 | ReLU | Same | No | No |
| Conv1D | 128 | 8 | ReLU | Same | No | No |
| Conv1D | 128 | 8 | ReLU | Same | No | No |
| Batch Norm | - | - | - | - | Yes (After Conv1D) | No |
| Dropout | - | - | - | - | No (After Batch Norm) | 0.6 |
| Conv1D | 64 | 8 | ReLU | Same | No | No |
| Conv1D | 64 | 8 | ReLU | Same | No | No |
| Flatten | - | - | - | - | No | No |
| Dense | - | - | ReLU | - | No | No |
| Dense | - | - | ReLU | - | No | No |
| Dropout | - | - | - | - | No (After Dense) | 0.6 |
| Dense (Output) | 7 | - | Softmax | - | No | No |

**Table 5:** CNN 1D model configuration.

| Classifier | WA | UA |
|---|---|---|
| KNeighborsClassifier | **62.94** | **56.72** |
| MLPClassifier | 62.39 | 54.72 |
| XGBClassifier | 48.61 | 44.07 |
| LGBClassifier | 33.01 | 42.30 |
| LogisticRegression | 50.82 | 41.58 |
| RandomForestCLassifier | 37.52 | 38.03 |
| GaussianNB | 23.89 | 32.74 |
| DecisionTreeClassifier | 19.38 | 22.92 |

**Table 6:** Baseline Accuracies (WA : Weighed Accuracy, UA : Unweighed Accuracy).

short-time blocks:

- The streaming or recorded speech was buffered into 1-s blocks to conduct block-by-block processing.
- The amplitude levels were normalized to the range -1 to 1.

There was a need to first buffer the speech files into 1-s blocks to conduct block-by-block processing.[3]

For each of the 1-s blocks of speech waveforms:

- A short-time Fourier transform was performed using a window of length '16' ms frames.
- The time-shift between subsequent frames was '4ms' giving 75% overlap between frames.
- The real and imaginary outputs from the short-time Fourier transform were converted to spectral magnitude values and concatenated across the whole frame to form an array.
- Resulting array was of shape 257 × 251 Where 257 is the number of frequency values (rows), and 251 is the number of time values (columns) in the spectrogram array.
- Spectral magnitude arrays of 257 × 251 real-valued numbers were converted into a color RGB image format.
- The transformation into the RGB format was based on the matplotlib "jet" colormap as they provided the best visual representation of speech spectrograms[18].

After the spectrogram images were saved onto the disk(Figure 5), a pre-trained CNN model called **MobilenetV2.0** was fine-tuned with raw spectrograms images (speech spectrogram arrays) as input.

4 different experiments were carried out by varying the **'overlapping window'** while segmenting the audio:

1.1. Image files were created by dividing input 'wav' files into blocks of 1 second *with overlapping windows of 10ms and including the Gender* of speakers.

---
[3]Directly calculating the spectrogram array for an audio 'wav' file of 3.5 seconds (avg length) will have 900 columns (sampled at 22.5KHZ) which is much more than the sequence length that can be fitted into a pre-trained model.

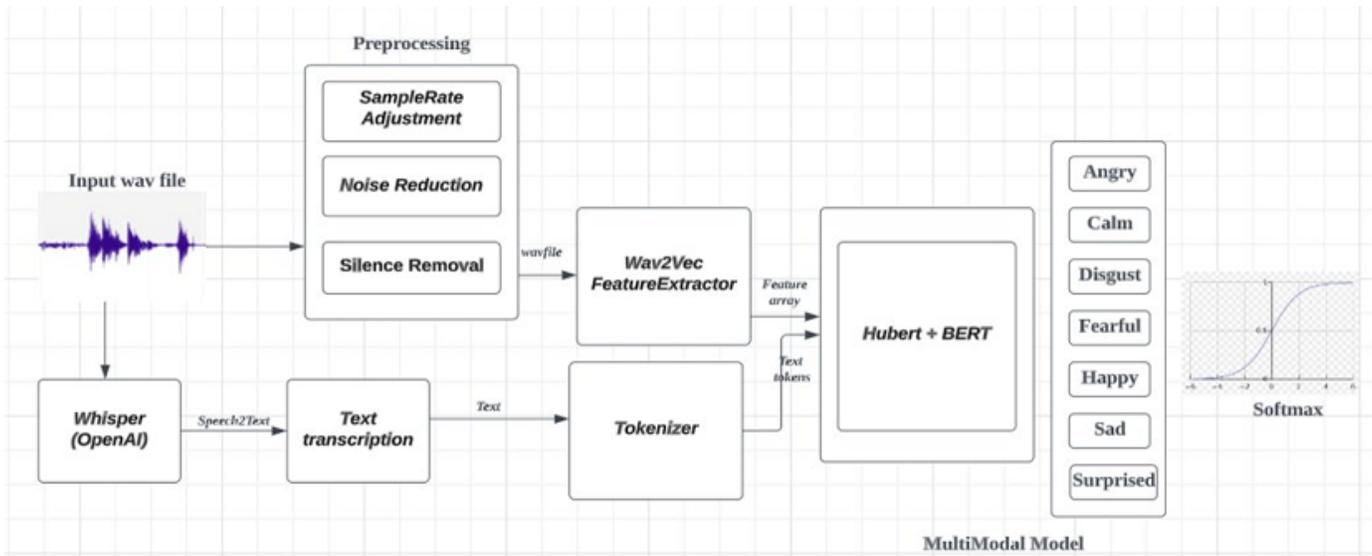

**Figure 6:** Inference from the proposed MultiModal model.

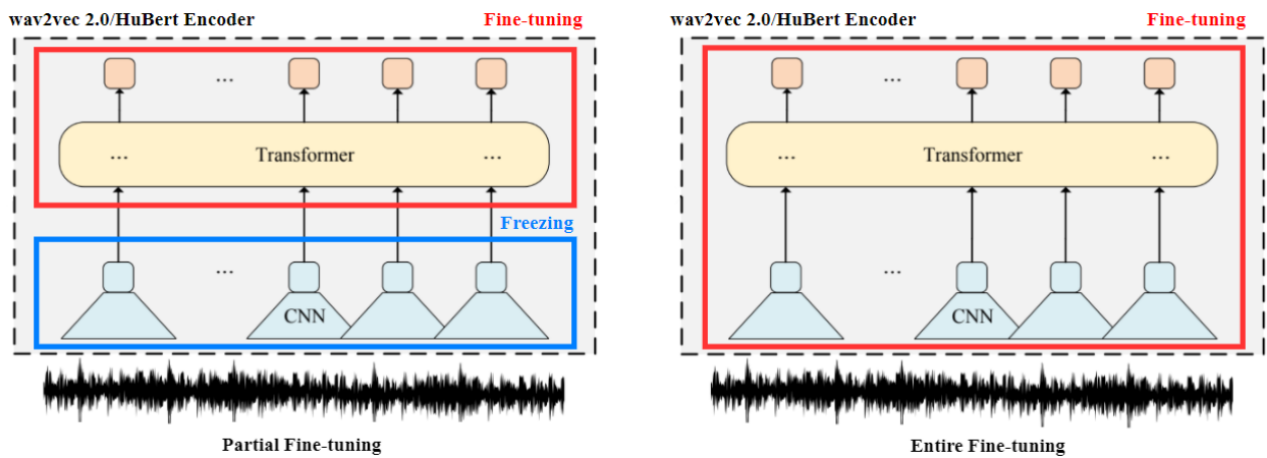

**Figure 7:** Partial fine-tuning (left) and entire fine-tuning (right) of wav2vec 2.0/HuBERT. Source:[39]

1.2. Image files were created by dividing input 'wav' files into blocks of 1 second *with overlapping windows of 10ms and excluding the Gender* of speakers.

1.3. Image files were created by dividing input 'wav' files into blocks of 1 second *without overlapping window of 10ms and including the Gender* of speakers.

1.4. Image files were created by dividing input 'wav' files into blocks of 1 second *without overlapping window of 10ms and excluding the Gender* of speakers.

For the experiments above where **'overlapping window'** was used, downsampling of 'wav' files was needed since otherwise, the number of images created would have been very high requiring substantially more computational power to process.

However, none of the experiments yielded satisfactory results, in fact, they could not even surpass the baseline. The result for the best-performing model among the above (1.4) is shown in table 8.

2. **CNN with 1D Convolution:**

Inspired by a published research[4], a CNN model was constructed. To prevent the model from overfitting, several techniques were employed, including the use of more aggressive dropout rates, L1 and L2 regularization, and batch normalization at different convolutions and fully connected layers.

The basic idea was The CNN would then learn to identify temporal patterns in the MFCCs and other features that are associated with specific emotions by applying 1D convolution.

---
[4]Intel Case Study

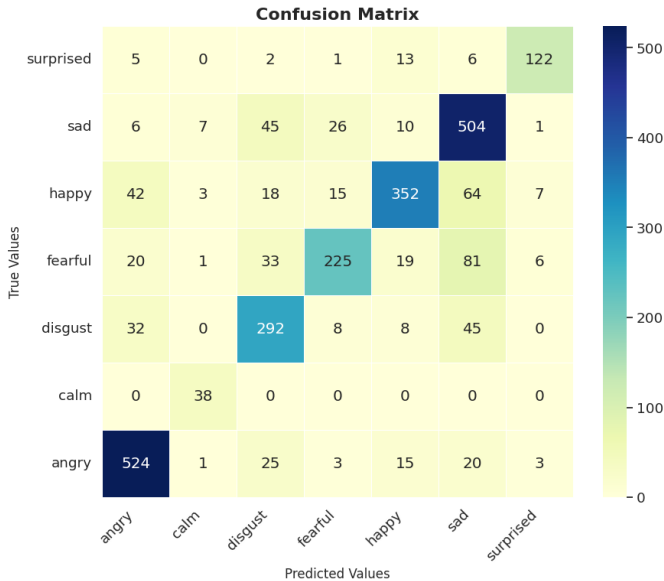

**Figure 8:** ConfusionMatrix for the test set

For instance, a CNN might learn to identify a pattern in the MFCCs that is associated with the emotion of happiness. This pattern might be a sequence of high values that correspond to the frequencies that are associated with happiness.

Model architecture is shown in the table 5. It makes use of 8 *Conv1D* layers, 3 *Dropout* layers, 2 *batchNormalization* layers and finally 3 *dense* layers. Once the CNN has learned to identify patterns in the MFCCs and other features, FFN (Feed Forward Layer) can be used to classify audio recordings into different emotions. In the architecture just described, the final *dense* layer acts as a Feed Forward layer.

The model was trained for 10 epochs using *'categorical cross entropy* loss.

Model's performance only slightly improved than the earlier fitted baselines.(Table 8)

*D. Transfer Learning (Hubert FineTuning):*

Hubert[13] employs a convolutional neural network (CNN) as its encoder. The CNN frontend performs feature extraction from the raw audio input.

It includes a transformer-based acoustic model. The transformer layer(s) in Hubert capture contextual information and allow the model to learn relationships between different parts of the input sequence. It helps in modeling long-range dependencies and generating contextualized representations.

Figure 7 illustrates the distinction between partial and entire fine-tuning on the wav2vec 2.0/HuBERT model. In the case of partial fine-tuning, the model is partitioned into two components: **CNN-based feature encoder** and a **transformer-based contextualized encoder**. During this process, the CNN-based encoder remains frozen, with its parameters fixed, while only the transformer blocks are fine-tuned. Partial fine-tuning operates as a form of domain adaptation at the top level, safeguarding the integrity of the lower CNN layers that possess inherent expressive capabilities.

Conversely, for complete fine-tuning, depicted on the right side of Figure 1, both the CNN and Transformer modules undergo fine-tuning throughout downstream training. Through training foundational features at the lower level, complete fine-tuning enables more comprehensive and targeted higher-level expressions. Assuming the fine-tuned wav2vec 2.0/HuBERT models already possess sufficient information-capturing power, we seamlessly augment them with straightforward downstream adaptors (e.g., classifiers/decoders) without introducing an additional burdensome and redundant encoder.

In the context of Speech Emotion Recognition (SER), a linear layer is directly appended as a downstream classifier. This linear layer serves as a fundamental component for utterance-level classification, with the overarching goal of minimizing cross-entropy loss.

Considering this recent success of Hubert model in speech recognition tasks, it was decided to finetune it for SER task. Experiments were done with 2 model architectures:

1. **Standalone Fine-tuned Hubert Model:**

    Using PyTorch as the backend, fine-tuning was performed. Initial experiments were done by unfreezing the top 2 CNN encoder layers of the Hubert model.

    However, it was observed that the model performs better without unfreezing any of the pre-trained Hubert layers. Hence, finally, weights were updated only for the classifier layer (on top of the Hubert model).

    The training process is described below:

    - Audio wav files from the train and validation datasets are resampled to a sampling rate of '16KHZ'. The resulting audio array is stored as a numpy array.
    - Using hugging face's feature extractor, specific features - MFCCs and their derivatives were extracted from the audio data stored in the array created in the earlier step.
    - Furthermore, these inputs are standardized and converted into a dataset object with key names of *input_values* and *labels* that can be used for training a model.

    Table 7 provides the values of network tuning parameters.

2. **Fine-tuned Hubert and BERT MultiModal Model:**

To improve the accuracy further, a 'Multi-Modal' architecture was developed that combines the strengths of both the Hubert and BERT models. Our model takes as input both audio features and text embeddings and concatenates the last hidden states from the Hubert and BERT models. The concatenated output is then passed through a linear classifier to produce the final predictions.

| Parameter | Parameter Value |
|---|---|
| Optimization Algorithm | AdamW |
| Max Epochs | 6 |
| Train Batch Size | 4 |
| Validation Batch Size | 4 |
| Gradient Accumulation Steps | 4 |
| Weight Decay | 0.01 |
| Warmup Steps | 500 |
| Learning rate | 5e-5 |

**Table 7:** FineTuning parameters for hubert-base-ls960

The Hubert model is used to process the audio input, while the BERT model is used to process the text input. Both models are pre-trained on large amounts of data and fine-tuned for our specific task. The last hidden states from both models are extracted, mean-pooled along the sequence dimension, and passed through a dropout layer for regularization. The resulting outputs are then concatenated along the feature dimension to achieve state-of-the-art performance on our task.

| Classifier | WA | UA |
|---|---|---|
| KNeighborsClassifier | 62.94 | 56.72 |
| MLPClassifier | 62.39 | 54.72 |
| XGBClassifier | 48.61 | 44.07 |
| MobilenetV2.0 | 38.03 | 37.52 |
| CNN (1D) | 63.10 | 56.70 |
| Hubert Standalone | 75.52 | 74.54 |
| **Hubert and BERT MultiModal** | **79.89** | **77.68** |

**Table 8:** Model's Test results (WA : Weighed Accuracy, UA : Unweighed Accuracy).

The training process is described below:

- The Hubert model which was fine-tuned earlier was utilized again.
- The text transcriptions were extracted for each audio file using Open-AI's Whisper model[5].
- During training, Input wav files were passed through to the Hubert model. At the same time, tokenized text transcriptions of the audio wav files were forwarded to the BERT model.

This way the MultiModal model is jointly trained on Text Embeddings and Wav file Embeddings on the complete Train dataset and evaluated on validation set.

Figure 6 illustrates the inference mechanism from this Model.

- When a 'wav' file is given as input to the Model, it is fed both into the Preprocessing pipeline as well as the OpenAI Whisper model.

[5] https://openai.com/blog/whisper/

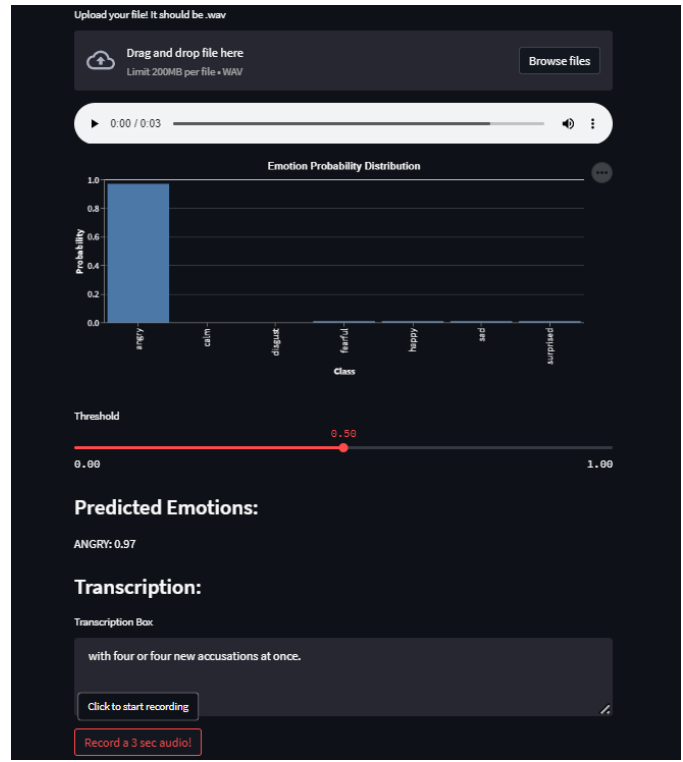

**Figure 9:** Web Application screenshot for the proposed deployed model.

- OpenAI's Whisper model does *SpeechToText* conversion to generate 'Text Transcriptions'.
- At the same time audio is pre-processed by applying *'SampleRate Adjustment'*, *'Noise Reduction'*, *'Silence Removal'*.
- Text transcriptions generated before are tokenized using BERT's tokenizer and simultaneously, the feature array is extracted from an earlier pre-processed audio file using the **Wav2VecFeatureExtractor**,
- Finally, both of these are given as input to the **MultiModal model** which gives output probabilities for each emotion class as well as text transcriptions as the final output.

| Dataset | WA | UA |
|---|---|---|
| RAVDESS | 78.45 | 79.31 |

**Table 9:** Benchmark test results on RAVDESS dataset

## IV. Results

Table 8 presents the overall results for different models. We use the standard performance metrics for accessing different models: sample weighted accuracy, **WA**, which corresponds to the percentage of correctly classified samples over all samples; unweighed (or balanced) accuracy, **UA**, which is the average of the individual class accuracies and is not affected by imbalanced classes.

All models were trained on the same train-validation-test splits and the table 8 captures the accuracies on the 'test set'.

Figure 8 shows the Confusion Matrix for the Multi-model on the 'test dataset'. One can see that the model very accurately identifies most of the 'angry' and 'sad' emotions. All 'calm' emotions are also identified without any error. However, the model does mix up 'sad' and 'happy' , 'angry' and 'happy' sometimes.

From the results, it is evident that the *'MultiModal architecture'* significantly outperforms others.

**Benchmarking and comparison**  Benchmark test was done with *RAVDESS* dataset [24] and it was found that our model achieved close to 80% accuracy on it. RAVDESS dataset was chosen as it was closest to the actual train dataset on which the model was trained and it requires only removal of the 'neutral' emotion label. Table 9 captures the benchmark test results.

Based on the comparison of our results presented in Table 8, we can conclude that the emotion recognition task benefits from the integration of different types of data. Our multimodal approach outperforms our speech-only test results and also performs better than other models.

We attribute our Model's strong performance to curating a robust training dataset (by combining multiple publicly available datasets), preprocessing and augmentation techniques, and the fact that we jointly train the powerful pre-trained BERT and Hubert transformer models.

## V. Web application

To test the model's performance seamlessly and on real-world speech snippets, a Web-based application was created using Streamlit [6] framework. The web application interface was kept simple and uncluttered. A provision was made to upload the audio files in 'wav' format. Also, a provision to record a 3 seconds clip was provided for someone to record their own short clips.

The model would then take the speech input, do the preprocessing, and invoke the Whisper model for transcription generation. Text tokenized outputs along with the speech features were passed together to the Multi-modal architecture as shown in the inference diagram (figure 6).

Figure 9 shows the graphical user interface along with the model's output consisting of the predicted Emotion probabilities and the transcripts.

## VI. Conclusion and future work

In this work, we evaluated different architectures for Speech emotion recognition tasks by training them on a combined dataset specifically curated using different audio augmentation and standardization techniques. We obtained state-of-the-art emotion recognition performance by including implicit linguistic information learned through joint finetuning of Hubert and BERT models. Our research shows that we are entering a new phase in the field of speech emotion recognition.

This phase is characterized by the use of pre-trained transformer-based models that provide a strong foundation for integrating the two main sources of information in spoken language: linguistics and paralinguistics. This integration has been long sought after and can be finally achieved.

In future work, we intend to extend our proposed model to use Hubert LARGE and X-LARGE versions for a more accurate speech model. For text, apart from Bidirectional Encoder Representations from Transformers (BERT), we plan to experiment with Generative Pre-training (GPT) and its variants, Transformer-XL, and Cross-lingual Language Models (XLM).

**Personality identification:**  Based on the findings from our work on speech emotion recognition, we intend to build a *personality detection* model to predict personality traits from the *Big-Five impressions*[25]. We also intend to explore the relation between emotions and personality traits and the possibility of predicting both of them using a *multi-task learning framework*.

## References

[1] Jo-Anne Bachorowski and Michael J. Owren. Vocal expression of emotion: Acoustic properties of speech are associated with emotional intensity and context. *Psychological Science*, 6(4):219–224, 1995.

[2] Abdul Malik Badshah, Jamil Ahmad, Nasir Rahim, and Sung Wook Baik. Speech emotion recognition from spectrograms with deep convolutional neural network. In *2017 International Conference on Platform Technology and Service (PlatCon)*, pages 1–5, 2017.

[3] Alexei Baevski, Yuhao Zhou, Abdelrahman Mohamed, and Michael Auli. wav2vec 2.0: A framework for self-supervised learning of speech representations. In H. Larochelle, M. Ranzato, R. Hadsell, M.F. Balcan, and H. Lin, editors, *Advances in Neural Information Processing Systems*, volume 33, pages 12449–12460. Curran Associates, Inc., 2020.

[4] Rishi Bommasani, Drew A. Hudson, and Ehsan Adeli et all. On the opportunities and risks of foundation models, 2022.

[5] Carlos Busso, Murtaza Bulut, Chi-Chun Lee, Ebrahim (Abe) Kazemzadeh, Emily Mower Provost, Samuel Kim, Jeannette N. Chang, Sungbok Lee, and Shrikanth S. Narayanan. Iemocap: interactive emotional dyadic motion capture database. *Language Resources and Evaluation*, 42:335–359, 2008.

[6] Houwei Cao, David G Cooper, Michael K Keutmann, Ruben C Gur, Ani Nenkova, and Ragini Verma. Crema-d: Crowd-sourced emotional multimodal actors dataset. *IEEE transactions on affective computing*, 5(4):377–390, 2014.

---
[6]https://streamlit.io/

[7] Ting Chen, Simon Kornblith, Mohammad Norouzi, and Geoffrey Hinton. A simple framework for contrastive learning of visual representations. In Hal Daumé III and Aarti Singh, editors, *Proceedings of the 37th International Conference on Machine Learning*, volume 119 of *Proceedings of Machine Learning Research*, pages 1597–1607. PMLR, 13–18 Jul 2020.

[8] Nicholas Cummins, Shahin Amiriparian, Gerhard Hagerer, Anton Batliner, Stefan Steidl, and Björn W. Schuller. An image-based deep spectrum feature representation for the recognition of emotional speech. In *Proceedings of the 25th ACM International Conference on Multimedia*, MM '17, page 478–484, New York, NY, USA, 2017. Association for Computing Machinery.

[9] Haytham M Fayek, Margaret Lech, and Lawrence Cavedon. Evaluating deep learning architectures for speech emotion recognition. *Neural Networks*, 92:60–68, 2017.

[10] H.M. Fayek, M. Lech, and L. Cavedon. Towards real-time speech emotion recognition using deep neural networks. In *2015 9th International Conference on Signal Processing and Communication Systems (ICSPCS)*, pages 1–5, 2015.

[11] Kun Han, Dong Yu, and Ivan Tashev. Speech emotion recognition using deep neural network and extreme learning machine. 09 2014.

[12] Ling He, Margaret Lech, Sheeraz Memon, and Nicholas Allen. Recognition of stress in speech using wavelet analysis and Teager energy operator. In *Proc. Interspeech 2008*, pages 605–608, 2008.

[13] Wei-Ning Hsu, Benjamin Bolte, Yao-Hung Hubert Tsai, Kushal Lakhotia, Ruslan Salakhutdinov, and Abdelrahman Mohamed. Hubert: Self-supervised speech representation learning by masked prediction of hidden units. *IEEE/ACM Trans. Audio, Speech and Lang. Proc.*, 29:3451–3460, oct 2021.

[14] Dias Issa, M. Fatih Demirci, and Adnan Yazici. Speech emotion recognition with deep convolutional neural networks. *Biomedical Signal Processing and Control*, 59:101894, 2020.

[15] Philip Jackson and Sana ul haq. Surrey audio-visual expressed emotion (savee) database, 04 2011.

[16] Sofia Kanwal and Sohail Asghar. Speech emotion recognition using clustering based ga-optimized feature set. *IEEE Access*, 9:125830–125842, 2021.

[17] Gil Keren and Björn Schuller. Convolutional rnn: an enhanced model for extracting features from sequential data, 2017.

[18] Margaret Lech, Melissa Stolar, Christopher Best, and Robert Bolia. Real-time speech emotion recognition using a pre-trained image classification network: Effects of bandwidth reduction and companding, May 2020.

[19] Y. LeCun, Fu Jie Huang, and L. Bottou. Learning methods for generic object recognition with invariance to pose and lighting. In *Proceedings of the 2004 IEEE Computer Society Conference on Computer Vision and Pattern Recognition, 2004. CVPR 2004.*, volume 2, pages II–104 Vol.2, 2004.

[20] Jinkyu Lee and Ivan J. Tashev. High-level feature representation using recurrent neural network for speech emotion recognition. In *Interspeech*, 2015.

[21] Dongge Li, Ishwar K. Sethi, Nevenka Dimitrova, and Tom McGee. Classification of general audio data for content-based retrieval. *Pattern Recognition Letters*, 22(5):533–544, 2001. Image/Video Indexing and Retrieval.

[22] Pengcheng Li, Yan Song, Ian Vince McLoughlin, Wu Guo, and Li-Rong Dai. An attention pooling based representation learning method for speech emotion recognition. In *Interspeech 2018*. International Speech Communication Association, September 2018.

[23] Wootaek Lim, Daeyoung Jang, and Taejin Lee. Speech emotion recognition using convolutional and recurrent neural networks. In *2016 Asia-Pacific Signal and Information Processing Association Annual Summit and Conference (APSIPA)*, pages 1–4, 2016.

[24] Steven R. Livingstone and Frank A. Russo. The Ryerson Audio-Visual Database of Emotional Speech and Song (RAVDESS), April 2018. Funding Information Natural Sciences and Engineering Research Council of Canada: 2012-341583 Hear the world research chair in music and emotional speech from Phonak.

[25] Robert R. McCrae and Oliver P. John. An introduction to the five-factor model and its applications. *Journal of personality*, 60 2:175–215, 1992.

[26] Annamaria Mesaros, Toni Heittola, Antti Eronen, and Tuomas Virtanen. Acoustic event detection in real-life recordings. 07 2014.

[27] Mustaqeem and Soonil Kwon. A cnn-assisted enhanced audio signal processing for speech emotion recognition. *Sensors*, 20(1):183, 2019.

[28] Sandra Ottl, Shahin Amiriparian, Maurice Gerczuk, Vincent Karas, and Björn Schuller. Group-level speech emotion recognition utilising deep spectrum features. In *Proceedings of the 2020 International Conference on Multimodal Interaction*, ICMI '20, page 821–826, New York, NY, USA, 2020. Association for Computing Machinery.

[29] Jiří Přibil and Anna Přibilová. An experiment with evaluation of emotional speech conversion by spectrograms. *Measurement Science Review*, 10:72–77, 03 2010.

[30] Aharon Satt, Shai Rozenberg, Ron Hoory, et al. Efficient emotion recognition from speech using deep learning on spectrograms. In *Interspeech*, pages 1089–1093, 2017.

[31] Klaus R. Scherer. Vocal affect expression: a review and a model for future research. *Psychological Bulletin*, 1986.

[32] Klaus R Scherer. Vocal communication of emotion: A review of research paradigms. *Speech Communication*, 40(1):227–256, 2003.

[33] Meenakshi Sood and Shruti Jain. Speech recognition employing mfcc and dynamic time warping algorithm. In Pradeep Kumar Singh, Zdzislaw Polkowski, Sudeep Tanwar, Sunil Kumar Pandey, Gheorghe Matei, and Daniela Pirvu, editors, *Innovations in Information and Communication Technologies (IICT-2020)*, pages 235–242, Cham, 2021. Springer International Publishing.

[34] Rui Sun, Elliot Moore, and Juan F. Torres. Investigating glottal parameters for differentiating emotional categories with similar prosodics. In *2009 IEEE International Conference on Acoustics, Speech and Signal Processing*, pages 4509–4512, 2009.

[35] Burak Uzkent, Buket D. Barkana, and Hakan Cevikalp. Non-speech environmental sound classification using svms with a new set of features. 2012.

[36] Ashish Vaswani, Noam Shazeer, Niki Parmar, Jakob Uszkoreit, Llion Jones, Aidan N. Gomez, Lukasz Kaiser, and Illia Polosukhin. Attention is all you need, 2023.

[37] Dimitrios Ververidis and Constantine Kotropoulos. Emotional speech recognition: Resources, features, and methods. *Speech Communication*, 48(9):1162–1181, 2006.

[38] Lode Vuegen, Bert Van Den Broeck, Peter Karsmakers, Jort F. Gemmeke, Bart Vanrumste, and Hugo Van hamme. An mfcc-gmm approach for event detection and classification. 2013.

[39] Yingzhi Wang, Abdelmoumene Boumadane, and Abdelwahab Heba. A fine-tuned wav2vec 2.0/hubert benchmark for speech emotion recognition, speaker verification and spoken language understanding, 2022.

[40] Mingke Xu, Fan Zhang, Xiaodong Cui, and Wei Zhang. Speech emotion recognition with multiscale area attention and data augmentation, 2021.

[41] Mingke Xu, Fan Zhang, and Samee U. Khan. Improve accuracy of speech emotion recognition with attention head fusion. In *2020 10th Annual Computing and Communication Workshop and Conference (CCWC)*, pages 1058–1064, 2020.

[42] Promod Yenigalla, Abhay Kumar, Suraj Tripathi, Chirag Singh, Sibsambhu Kar, and Jithendra Vepa. Speech emotion recognition using spectrogram & phoneme embedding. In *Interspeech*, volume 2018, pages 3688–3692, 2018.

[43] Yuanyuan Zhang, Jun Du, Zirui Wang, Jianshu Zhang, and Yanhui Tu. Attention based fully convolutional network for speech emotion recognition. In *2018 Asia-Pacific Signal and Information Processing Association Annual Summit and Conference (APSIPA ASC)*. IEEE, nov 2018.